\documentclass[letterpaper, 10 pt, conference]{ieeeconf}  

\IEEEoverridecommandlockouts                              

\usepackage{graphics} 
\usepackage{times} 
\usepackage{amsmath} 
\usepackage{amssymb}  
\usepackage{algorithmic}
\usepackage{graphicx}
\usepackage{textcomp}
\usepackage[usenames, dvipsnames, table]{xcolor}
\usepackage{float}
\usepackage[style=ieee,sorting=none,backend=biber, maxbibnames=99]{biblatex}
\usepackage{hyperref}
\usepackage[ruled,linesnumbered,resetcount,vlined]{algorithm2e}
\usepackage{commath}
\usepackage{caption}
\usepackage{subcaption}
\usepackage{mathtools}

\usepackage{array}

\usepackage{bbding}
\usepackage{pifont}
\usepackage{wasysym}
\usepackage{xcolor}

\let\labelindent\relax
\usepackage[inline]{enumitem}

\colorlet{mylinkcolor}{BrickRed}
\colorlet{mycitecolor}{Green}
\colorlet{myurlcolor}{NavyBlue}

\hypersetup{
  linkcolor  = mylinkcolor,
  citecolor  = mycitecolor,
  urlcolor   = myurlcolor,
  colorlinks = true,
}

\usepackage[capitalize]{cleveref}

\title{\LARGE \bf Clutter-Aware Spill-Free Liquid Transport via Learned Dynamics}

\author{Ava Abderezaei$^{*}$, Anuj Pasricha, Alex Klausenstock, and Alessandro Roncone%
\thanks{$^{*}$ Corresponding author.}
\thanks{All authors are with the Department of Computer Science, University of Colorado Boulder, 1111 Engineering Drive, Boulder, CO USA. This work is partly supported by NSF \#2222952/2953.
{\tt\small firstname.lastname@colorado.edu}}
}

\begin{document}
\maketitle
\thispagestyle{empty}
\pagestyle{empty}

\begin{abstract}
In this work, we present a novel algorithm to perform spill-free handling of open-top liquid-filled containers that operates in cluttered environments.
By allowing liquid-filled containers to be tilted at higher angles and enabling motion along all axes of end-effector orientation, our work extends the reachable space and enhances maneuverability around obstacles, broadening the range of feasible scenarios.
Our key contributions include: i) generating spill-free paths through the use of RRT* with an informed sampler that leverages container properties to avoid spill-inducing states (such as an upside-down container), ii) parameterizing the resulting path to generate spill-free trajectories through the implementation of a time parameterization algorithm, coupled with a transformer-based machine-learning model capable of classifying trajectories as spill-free or not.
We validate our approach in real-world, obstacle-rich task settings using containers of various shapes and fill levels and demonstrate an extended solution space that is at least 3x larger than an existing approach.

\end{abstract}
\section{Introduction}
The integration of robots is rapidly expanding across various domains, spanning from industrial automation to healthcare, agriculture, and domestic services. The ability to manipulate objects is fundamental for robots to perform a variety of tasks in these diverse domains \cite{technologies9010008, doi:10.1146/annurev-control-060117-104848}. Among the different categories of objects that robots encounter, liquids present a unique challenge due to their complex behavior. Despite the importance of robotic fluid manipulation in various applications such as handling hazardous materials in laboratories to facilitating assistive feeding technologies designed for differently-abled individuals, research in robotic fluid manipulation is under-explored and challenging for several reasons.
\begin{figure}
    \centering
    \includegraphics[width=\columnwidth]{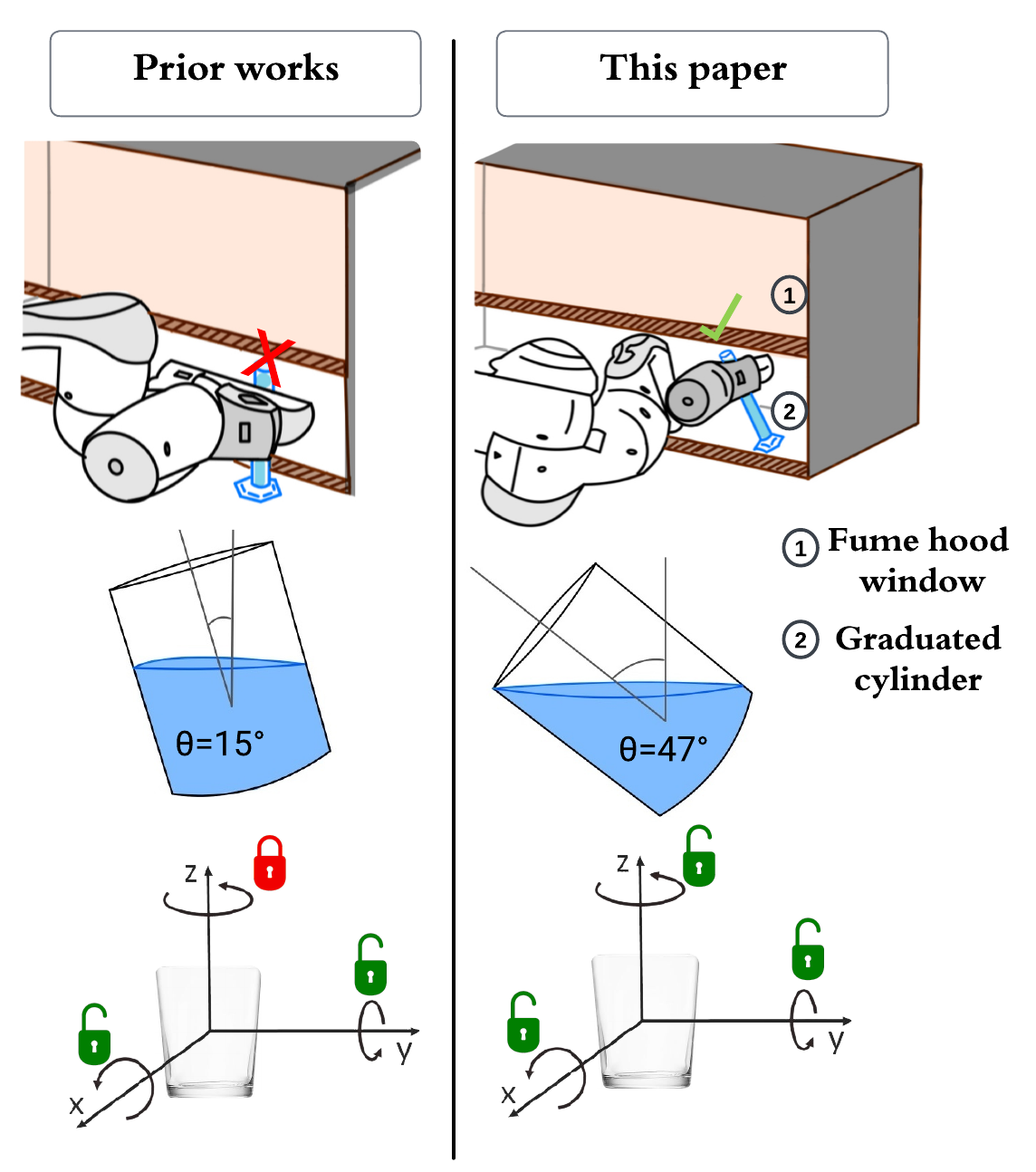}
    \caption{
    Motivated by the need for robots to perform in cluttered environments such as a chemistry setup where robots must transport chemistry vessels inside a fume hood (\textsl{top row}), 
    this paper targets the challenge of enabling spill-free liquid transport while avoiding obstacles. Our approach introduces two key enhancements over existing methods: (i) increasing the allowable tilt angle to the maximum quasi-static spill-free tilt angle, rather than the limited angles used previously \cite{pendulum, pendulum2019} (\textsl{second row}); (ii) enabling rotation about all axes throughout the trajectory, instead of maintaining fixed yaw or orientation (\textsl{third row}) \cite{pendulum,  springdamper2021}.
    }
    \label{fig:first_page_figure}
\end{figure}

Fluid manipulation is challenging due to the intricate and nonlinear nature of fluid dynamics. Accurately modeling fluid dynamics is heavily task-dependent and requires substantial computational resources, hindering real-time control \cite{systemstudy1, systemstudy2}. Existing approaches tackle these challenges by simplifying the complex fluid dynamics into more computationally efficient linearized pendulum or mass-spring-damper models. However, these methods struggle in cluttered scenes due to several reasons. These linearized models restrict the motions to pendulum-like or mass-spring-damper-like behavior \cite{pendulum2, pendulum2018, pendulum, springdamper2022, springdamper2021, pendulum2019}. Additionally, these approaches impose restrictions on the container's orientation by setting fixed or limited Euler angle ranges throughout the trajectory \cite{pendulum, springdamper2021}. Moreover, these methods rely on constant monitoring of the liquid surface to execute trajectories, which can be obstructed by opaque fumes produced by the liquid or by obstacles present in cluttered scenes \cite{spill1, pendulum2019,montazeri2013chemical}. Consequently, the action space is over-constrained, and the robot can only perform limited motions, inhibiting its ability to maneuver around obstacles in cluttered environments.

To address these limitations, we propose Spill-Free RRT* (\textit{SFRRT*})\footnote{Open-source code and video at: \href{https://avv-va.github.io/sfrrt}{https://avv-va.github.io/sfrrt}.}, for spill-free manipulation of open-top liquid-filled containers, particularly in cluttered environments. Our method enables these advantages (\cref{fig:first_page_figure}):
(i) \textsl{Increased action space}, so that the robot can operate effectively in cluttered scenarios. We do this by training a model to implicitly learn liquid dynamics to capture the correlation between container trajectory and spillage risk, 
(ii) \textsl{No additional information needed}. Our method relies solely on the robot and the container shape to operate and does not require constant monitoring of the liquid surface.

To achieve this, SFRRT* generates spill-free trajectories through these key contributions: (i) it first uses the container shape to inform the generation of a spill-free path via RRT*. By considering the container geometry, the algorithm calculates the maximum tilt angle beyond which the container will spill and constrains the path search accordingly (\cref{subsec:informed-geometric-path-planner}). (ii) Subsequently, our method generates a spill-free trajectory from this path by using a time-parameterization algorithm in conjunction with a transformer-based neural network classifier. The neural network, trained on a dataset of container trajectories, predicts whether a given trajectory is spill-free or not, allowing SFRRT* to set the trajectory's jerk to avoid spills (\cref{method:timepar}), (iii) Lastly, to train the transformer-based neural network, we produce a dataset of 700 container trajectories collected from individuals handling different container scenarios.

\section{Related Work}
Transporting open-top, liquid-filled containers requires the modeling of liquid motion along with the planning of time-parameterized motion paths.
A common approach in liquid modeling relies on using Computational Fluid Dynamics (CFD) simulations to achieve spill-free motions by designing controllers to suppress or limit slosh effects, i.e. liquid vibration effects \cite{cfdbook1, cfdbook2}.
Despite the accuracy of CFD simulations, they are computationally demanding and require considerable time to converge, making real-time control applications in robotics unfeasible \cite{cfd1, cfd2, meal1}. 

To overcome the computational cost associated with CFD, prior work considers two approaches for modeling the sloshing dynamics inside a container: a spherical pendulum \cite{pendulum, spill1, pendulum2019} and a two Degrees-of-Freedom (2DoF) mass-spring-damper system \cite{springdamper2021, springdamper2022}. 
These models are inherently nonlinear and computational constraints necessitate linearization for practical implementation. However, linearization introduces several kinematic constraints that restrict robots' range of motions, thereby limiting their ability to navigate through cluttered scenes.

\textit{Spherical pendulums.} These approaches present controllers that are designed to prevent liquid spillage \cite{spill1, pendulum2019}.
They limit the cup's freedom to only one axis of orientation, while also relying on continuous sensor input to monitor the liquid surface for modeling slosh dynamics. In contrast, our approach dispenses with the need for continuous liquid surface observation and allows the container to retain all its degrees of freedom.
Additional work assumes that a higher-level motion planner already provides an obstacle-free trajectory \cite{pendulum}. A quadratic program is then employed to optimize the trajectory within a linearized spherical pendulum dynamics model. While the method eliminates the need for continuous liquid surface monitoring, the linearization restricts object yaw to be constant and allows only tilting the container up to 15$^{\circ}$ to maintain acceptable linearization accuracy. In contrast, our approach enables motion along all three rotation axes and allows the container to be tilted to significantly larger angles, up to the point of spillage, by modeling the quasi-static fluid dynamics.

\textit{Mass-spring-dampers.} In these approaches, a sloshing-height evaluation method is employed for time-optimal trajectory planning \cite{springdamper2021}. However, this method is tailored for cylindrical containers subject to 1D and 2D motions. This sloshing estimation technique is extended to accommodate 3D motions but the model assumes that the container remains upright during motion \cite{springdamper2022}. In contrast, our approach allows the container to be tilted on each axis and is compatible with various container shapes.

Finally, a notable limitation of both mass-spring-dampers and pendulum-based slosh dynamics models is their inability to navigate around obstacles consistently. These models rely on maintaining pendulum-like or mass-spring-damper-like motions for spill-free movement, and obstacles can disrupt these motions, rendering these approaches ineffective.

\textit{Other Approaches.} 
Among other approaches, \cite{gompfit} presents an effective method for the rapid, spill-free transport of containers. However, it explicitly states that this technique is not intended for the transportation of liquids, as it does not incorporate constraints to inhibit the accumulation of energy within the cup, which could subsequently lead to sloshing.
Additionally, the method proposed in \cite{sloshbydata} aims to accurately learn slosh dynamics through machine learning, taking into account various liquid properties including thermodynamics. However, its applicability to motion planning with robots has not been evaluated, and its practical implementation necessitates continuous accurate liquid surface monitoring.

Overall, our approach outperforms prior methods by implicitly learning the dynamics of liquid spillage. It does not impose kinematic constraints that would limit the robot's range of motion such as setting a fixed container orientation, thereby enabling its maneuverability in cluttered environments. Moreover, it does not require constant observation of the liquid surface to generate motions and it is not restricted to a particular container shape.
\section{Methods}
\begin{figure*}
    \centering
    \includegraphics[width=\textwidth]{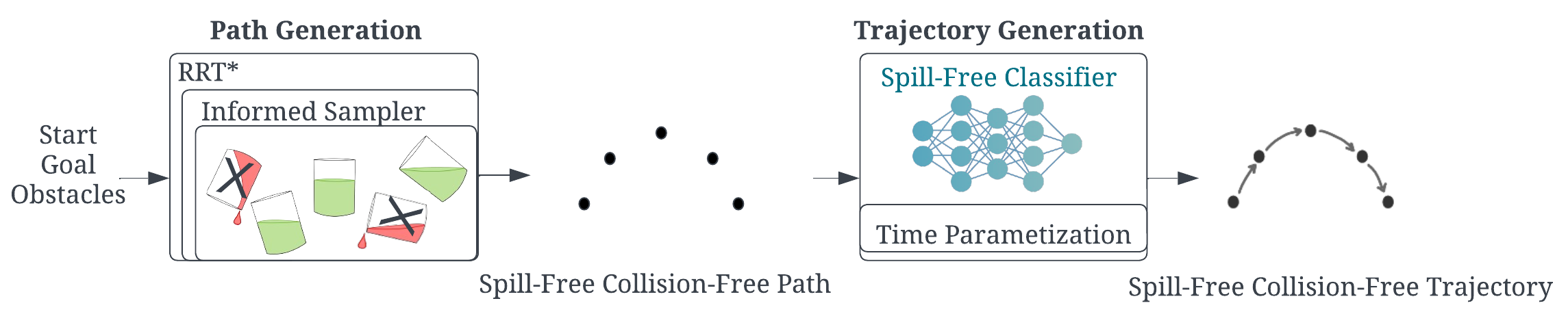}
    \caption{\textbf{System diagram}.
    The inputs are the start, goal, and obstacles, and the output is the trajectory. First, (i) a spill- and collision-free path is generated using RRT* and the informed sampler. The informed sampler utilizes the container properties to avoid sampling states such as an upside-down container. Then, (ii) a spill- and collision-free trajectory is generated using a time parameterization algorithm alongside the spill-free classifier model. The SFC model classifies trajectories as either spill-free or not and is utilized to set the jerk of the trajectory, ensuring the generation of a spill-free trajectory.}
    \label{fig:system_diagram}
    \vspace{-15pt}
\end{figure*}

This section introduces SFRRT*, a motion planner that generates spill- and collision-free trajectories (\cref{fig:system_diagram}).
SFRRT* accomplishes this through two key contributions: a) the informed sampler, mathematically formalized based on container geometry, to efficiently explore the space of spill-free robot states (\cref{subsec:informed-geometric-path-planner}), b) a transformer-based classifier that models the underlying nonlinear dynamics of liquid motion and classifies whether a planned trajectory is spill-free or not (\cref{method:timepar}).
\subsection{Mathematical Formulation of the Spillage Scenario}
\label{subsec:cup_max_tilt_angle}
A challenge in spill-free trajectory generation is that the volume of the robot configuration space that contains spill-free trajectories is small and therefore inefficient to explore with random sampling alone \cite{zach}. This dictates the need for an informed approach to explore dynamically-valid robot states. However, developing a model for combined robot and object dynamics, specifically spillage dynamics, is computationally expensive, task-specific, and impractical for real-world control.
To address this, we need to mathematically formalize how container properties inform the construction of the \textit{spill-free space}. For example, for a given container that contains a greater volume of liquid, it cannot be tilted as much as when it contains less liquid. Additionally, in scenarios where containers have the same liquid volume, taller containers can be tilted more than shorter ones.

\begin{figure*}
\centering
\begin{subfigure}{0.32\textwidth}
\includegraphics[width=\linewidth]{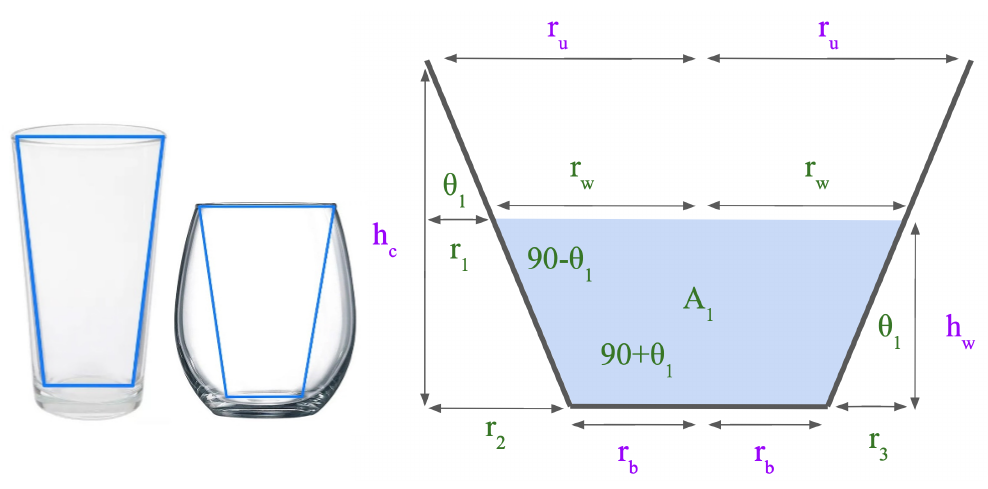}
\caption{Resting Position}
\label{fig:cup_math_rest}
\end{subfigure}
\hfill
\begin{subfigure}{0.30\textwidth}
\includegraphics[width=\linewidth]{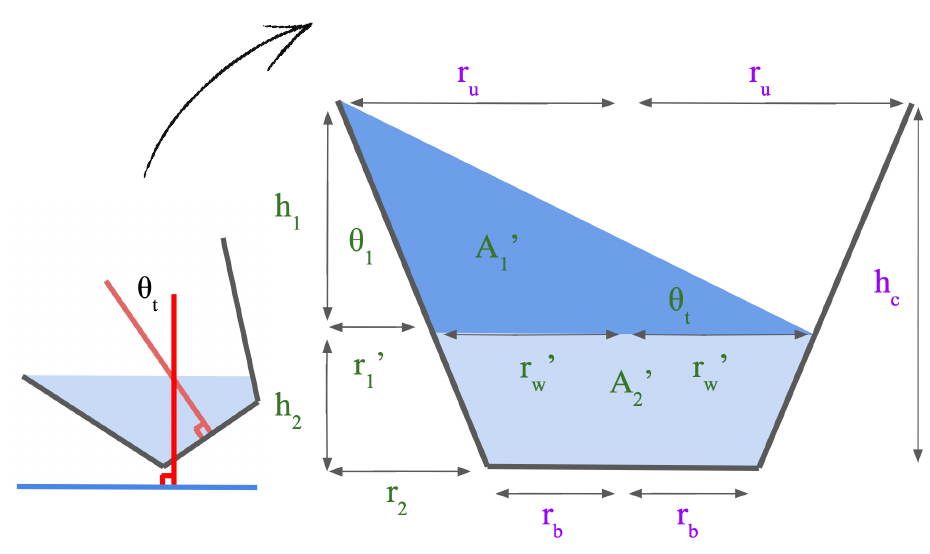}
\caption{Case I}
\label{fig:cup_math_case_1}
\end{subfigure}
\hfill
\begin{subfigure}{0.30\textwidth}
\includegraphics[width=\linewidth]{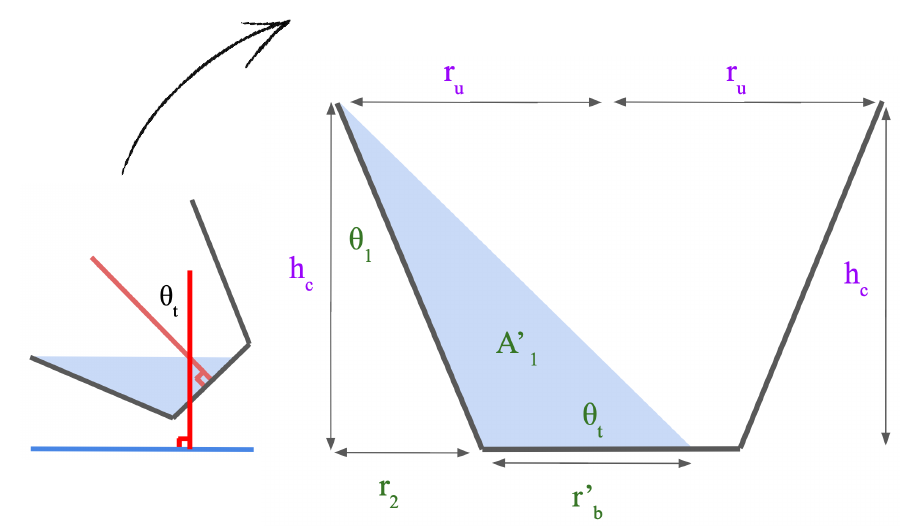}
\caption{Case II}
\label{fig:cup_math_case_2}
\end{subfigure}
\caption{(a) depicts modeling containers as frustum of cones. Two cases occur when tilting the container as illustrated in (b) and (c). Case I is when the liquid forms both a trapezoid and a triangle shape. Case II shows when the liquid forms only a triangular shape. In both cases, $\theta_t$ represents the maximum tilt angle being calculated.
}
\label{fig:cup math}
\end{figure*}

To sample spill-free configurations from the \textit{spill-free space}, we first calculate the maximum tilt angle that a container can withstand without spilling. Then, we sample states ensuring that their orientation remains within this range. Specifically, under quasi-static conditions, we consider the scenario that involves tilting a container to the point where it cannot retain all its content, i.e., to the angle $\theta_{max}$, such that any $\theta_{t} \leq \theta_{max}$ would be spill-free and any $\theta_t > \theta_{max}$ would cause spillage (\cref{fig:cup math}). Quasi-static conditions are assumed because they simplify analytical modeling without requiring expensive computational fluid dynamics calculations.

To formalize the spillage scenario, we model the container as a frustum of a cone, which accurately represents most common container shapes and can provide an upper bound on the maximum tilt angle of containers with other shapes, such as round bottom flasks or curvy wine glasses (\cref{fig:cup_math_rest}).
Specifically, we simplify curved containers to the largest possible frustums of cones that can fit inside the original shape, leading to predicting spillage at a lesser tilt angle than the real container's capabilities. This guarantees a $\theta$ less than $\theta_{max}$, ensuring the generation of spill-free states. Lastly, we simplify the frustum of the cone representation from 3D to 2D, resulting in a trapezoid. This simplification also allows us to represent planar symmetric containers such as rectangular prisms and can be directly extended to 3D. 

To solve for the maximum tilt angle, we assume that the information we have from the container is the bottom circle radius $r_b$, the top circle radius $r_u$, the cup height $h_c$, and the water height $h_w$ (\cref{fig:cup_math_rest}). With this information, we can derive other relevant variables for calculating the maximum tilt angle. As shown in \cref{fig:cup math}, two spillage scenarios could occur: 
\begin{enumerate*}[label=(\roman*)]
\item one where the liquid forms a trapezoid and a triangle (\cref{fig:cup_math_case_1}), and
\item another where the liquid only forms a triangle (\cref{fig:cup_math_case_2}).
\end{enumerate*} This happens because static liquid stays parallel to the horizontal ground and depending on the liquid level, these cases can occur.

To calculate the maximum tilt angle, for both spillage cases, we leverage the key insight that the total amount of liquid remains constant before and after tilting the container. By equating the liquid area in the initial and tilted configurations, we derive an algebraic formula to compute the angle. The resulting equations, \cref{eq:caseI} for case I and \cref{eq:caseII} for case II, provide the respective maximum tilt angles.

Maximum tilt angle for case I:
\begin{equation}
\theta_{max} = tan^{-1} \left[\frac
{h_c(r_u-r_{w'})}
{(r_u-r_b)(r_u+r_{w'})}\right]
\label{eq:caseI}
\end{equation}
where $r_{w'}= \frac{h_cr_ur_b - h_cr_2r_b + h_wr_2r_w + h_wr_2r_b}{h_cr_2 + h_cr_b}$.

Maximum tilt angle for case II:
\begin{equation}
\theta_{max} =tan^{-1} \left[ \frac
{h_c}
{r_u-r_b+(\frac{2h_w(r_w+r_b)}{h_c})}\right]
\label{eq:caseII}
\end{equation}
where $r_{w}= r_u-[(\frac{r_u-r_b}{h_c})(h_c-h_w)]$.

The next section explains how these equations are incorporated in the sampling-based path planner to inform the construction of a spill-free path.

\subsection{Informed Geometric Path Planning for Spill-Free Motions}
\label{subsec:informed-geometric-path-planner}
Having formalized the maximum tilt angle $\theta_{max}$, it is now used to guide the exploration of states that avoid spills, such as an upside-down container, or a container that is tilted beyond its capacity (i.e., $>$ $\theta_{max}$).
To achieve this, we implemented an informed sampler to sample robot configurations where the container orientation is less than the spill threshold $\theta_{max}$. Orientation refers to the container's Euler angles, and it is sampled such that the angle between the vertical axis and the axis about which the container is symmetric, is less than $\theta_{max}$. This informed sampler is then incorporated into a geometric RRT* path planner, to generate spill-free paths. RRT* is chosen due to its asymptotic optimality properties. By preferring transitions with minimal cost, RRT* generates smooth motion profiles that maintain orientation stability \cite{rrt*}. 
The next section describes how the path generated by RRT* is parameterized to ensure a spill-free trajectory.

\subsection{Time Parameterization To Create Spill-Free Trajectories}
\label{method:timepar}

While RRT* generates spill-free paths, to ensure the generation of a spill-free trajectory, we must carefully parameterize the path such that the velocity, acceleration, and jerk, do not cause spillage.
\begin{figure}
    \centering
\includegraphics[width=0.8\columnwidth]{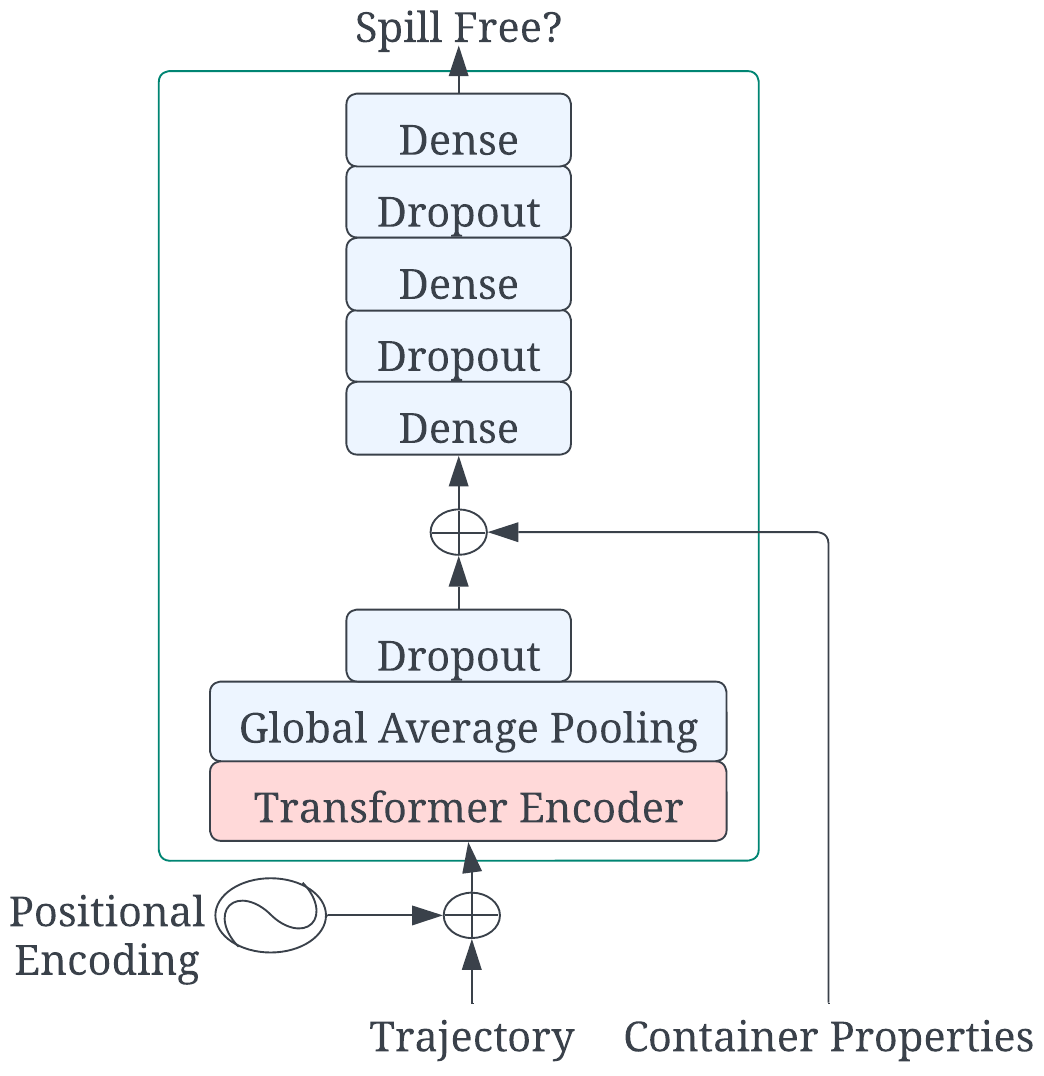}\caption{\textbf{Spill-Free Classifier Model Architecture} The input is the trajectory and the container properties. The output is a boolean stating whether the trajectory is spill-free or not.
}
    \label{fig:models}
\end{figure}
To do this, it is necessary to first model the risk of spillage with respect to the trajectory. To capture this relationship, we employed a transformer-based machine learning model trained to classify whether a container's trajectory is spill-free or not. This model is termed the \textit{Spill-Free Classifier} (SFC), and utilizes the transformer architecture, as depicted in \cref{fig:models}.

The input to SFC is the container trajectory and container properties, where the trajectory is the position and orientation over time and the properties represent the container's top and bottom radius, container height, and liquid fill level.
Furthermore, as shown in \cref{fig:models}, the trajectory is first passed through the transformer encoder layer. Then, the container properties are concatenated with it. The SFC model employs the transformer architecture because its input is sequence data (the trajectory) and transformers have become the leading architecture for handling sequential data due to their efficient parallel training capabilities and long context window \cite{huggingfacetransformers}. 

Lastly, to train the SFC model, we utilized a motion capture system to collect container trajectories (comprised of position and orientation over time) performed by individuals across different container types and liquid volumes (\cref{fig:Training set}). We collected a total of 700 trajectories, including 655 human-performed trajectories and 45 trajectories from the Panda robot. These trajectories encompassed a variety of paths from holding cups upside down and following zig-zag patterns to transporting cups in straight or typical paths between two points.
\begin{algorithm} \footnotesize
\caption{Spill Free Time Parametization (SFTP)} \label{alg:sftp}
\begin{algorithmic}[1]
\STATE {\bfseries Input:} Waypoints $\mathcal{W}$, Max Jerk $J_{max}$, Max Acceleration $A_{max}$, Max Velocity $V_{max}$, Jerk Decrease Factor $rate$, Max Iterations $N$,
\STATE {\bfseries Output:} Trajectory $\mathcal{T}$
\STATE $P_{s}, P_{f} \gets W[0]$, $W[length(W)-1]$
\STATE $V_{s}, V_{f}, A_{s}, A_{f} \gets 0$
\FOR{$ i = 1 ... N$} \label{sftp:loop}
    \STATE $\mathcal{T} \gets$ Ruckig($\mathcal{W}, J, V, A, P$)
    \STATE $\mathcal{T}_{fk} \gets ForwardKinematics(P)$ \label{sftp:fk}
    \STATE $spillFree \gets$ SpillFreeClassifier($\mathcal{T}_{fk}$) \label{sftp:sfc}
    \IF{$spillFree$}
        \RETURN $\mathcal{T}$ \label{sftp:stop}   
    \ELSE
    \STATE $J_{max} \gets \frac{J_{max}}{rate}$ \label{sftp:jerk_reduce}
    \ENDIF
\ENDFOR
\end{algorithmic}
\end{algorithm}

Having modeled the relationship between the container trajectory and spillage risk, we leverage SFC to generate spill-free trajectories. We introduce the \textit{Spill-Free Time Parameterization} (SFTP), which guarantees the creation of spill-free trajectories by utilizing a time-parameterization algorithm alongside the SFC model.
The key idea is to find the maximum jerk that does not cause spillage. To achieve this, SFTP works by iteratively (line \ref{sftp:loop}, \cref{alg:sftp}) reducing the maximum jerk of the trajectory (line \ref{sftp:jerk_reduce}, \cref{alg:sftp}), querying SFC after each iteration (line \ref{sftp:sfc}, \cref{alg:sftp}) until SFC predicts the resulting trajectory is spill-free. Jerk constraints are especially critical in preventing sudden changes in acceleration that could cause spillage. SFTP only modifies the jerk parameter since it directly affects acceleration and velocity. Lastly,  for time parameterization the Ruckig algorithm \cite{ruckig} is used since it can enforce constraints on higher-order derivatives including jerk, acceleration, and velocity.

In summary, to generate a spill-free collision-free trajectory (\cref{fig:system_diagram}), first, the container shape and fill level are leveraged to inform the generation of a path using RRT*. The spill-free classifier (SFC) model is then used alongside the Ruckig time-parameterization algorithm to ensure the generation of a spill-free trajectory. We refer to this combined process of generating the path and the trajectory, the \textit{spill-free RRT*} (SFRRT*) algorithm. 

\section{Evaluation}
\begin{figure}
\centering
\includegraphics[width=1\columnwidth]{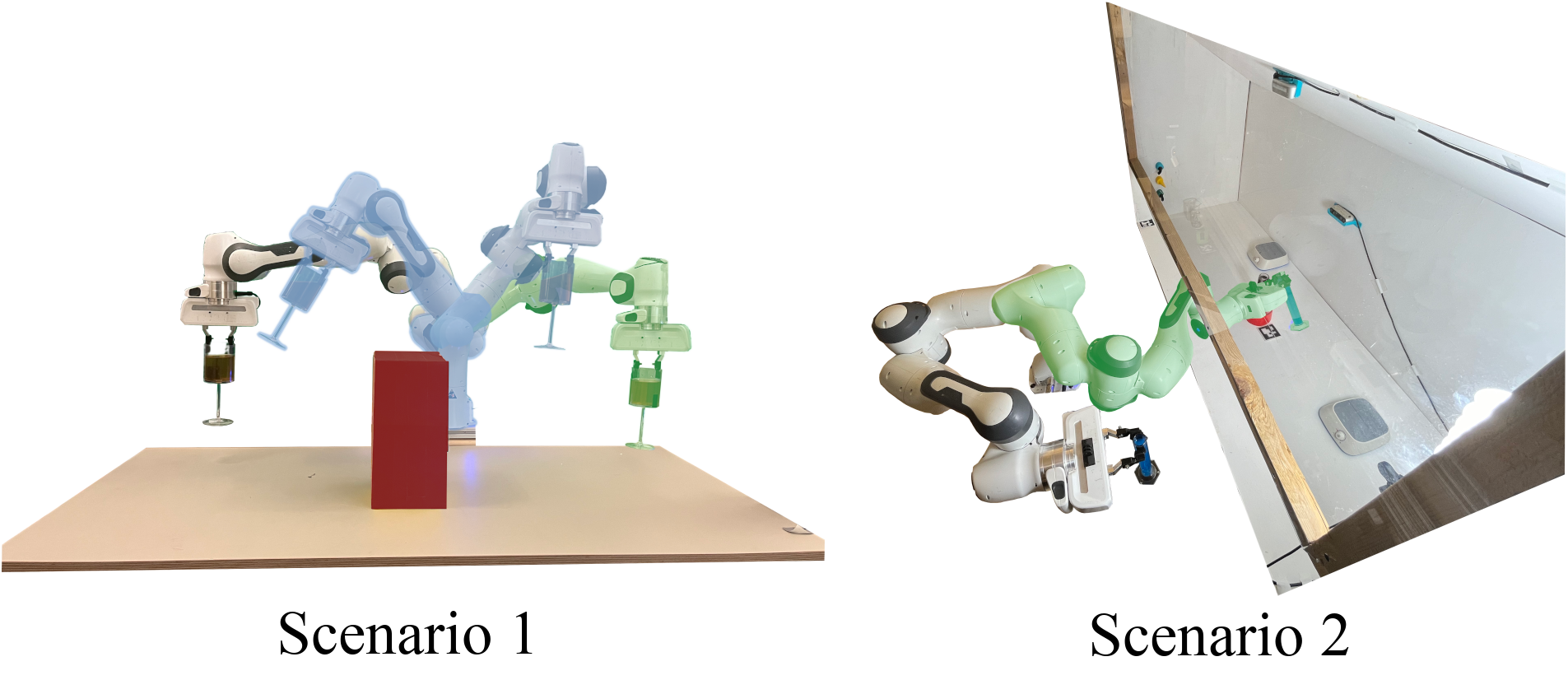}
\caption{
The two scenarios from \cref{table:tasks}: The original colored robot for start position, green for goal. Blue indicates intermediate robot states.
}
\label{fig:scenario_1_2}
\end{figure}

\begin{figure}
    \centering
\includegraphics[width=0.7\columnwidth]{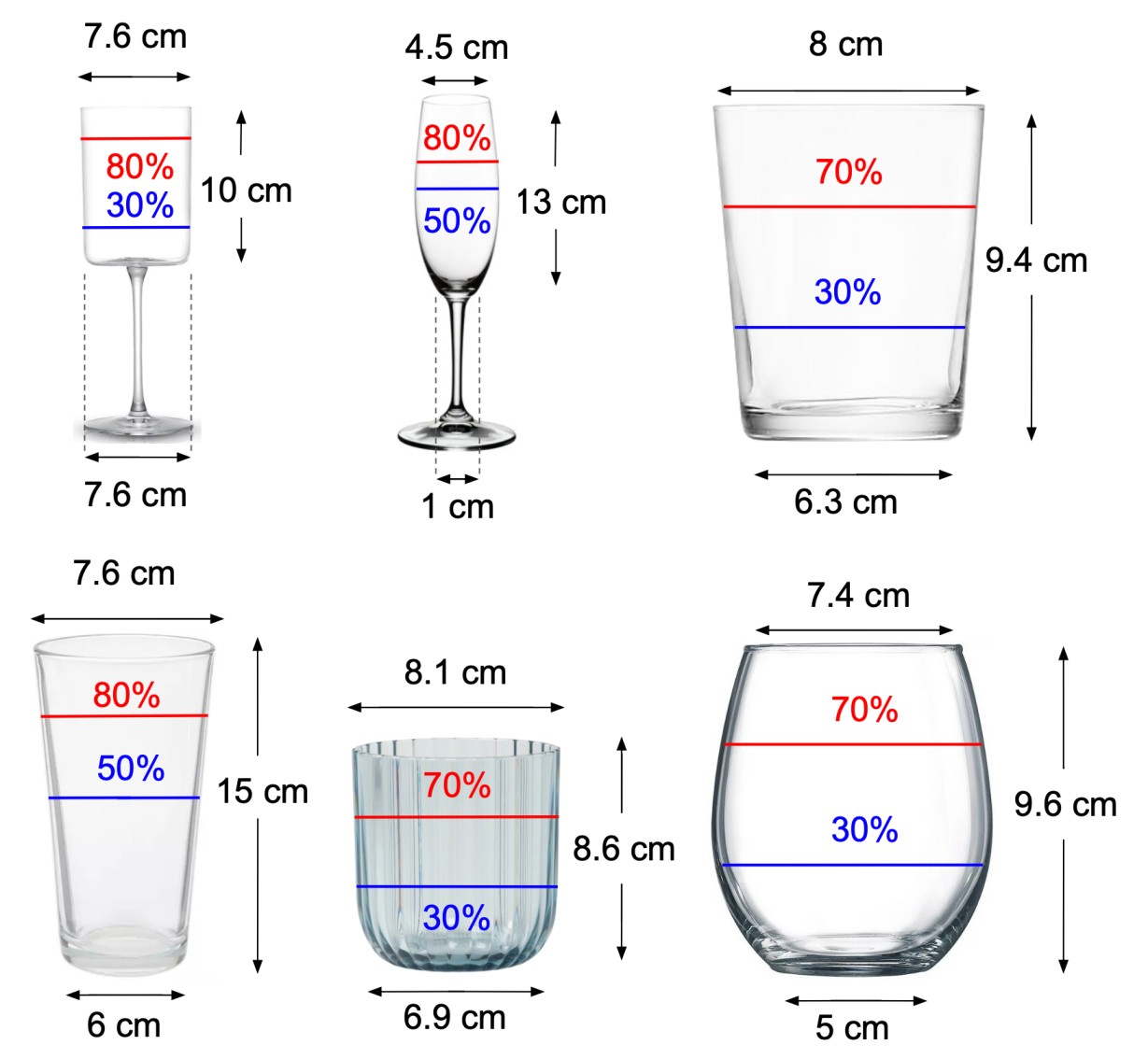}
    \caption{The entire set of containers used to train SFC. Notably, the top three containers, identified as the wine glass, flute glass, and basic glass, are used in our experiments.
    }
    \label{fig:Training set}
\end{figure}

We evaluate our spill-free collision-free trajectory planner, SFRRT*, on the 7DoF Franka Emika Panda robotic arm. We design two scenarios and utilize three different containers with different fill levels (\cref{fig:Training set}, top row) to quantify the performance of SFRRT*. These scenarios are designed to encapsulate real-world use cases of our motion planner, specifically cases that exhibit a diverse range of robot motions: Scenario 1 (\cref{fig:scenario_1_2}, left) involves the robot transporting a drink while avoiding the obstacles on the table; and, Scenario 2 (\cref{fig:scenario_1_2}, right) involves the robot in a chemistry lab transporting a container while avoiding collisions with the fume hood.
Additionally, the containers (wine, flute, and basic glasses) are picked based on their diverse geometries, while the fill levels, ranging from 30\% to 80\% are chosen to cover tasks with varying risks of spillage.
The objective across our evaluation tasks is to move these containers from the start to the goal states without spilling the contents of the container being transported. Each experiment, as enumerated in \cref{table:tasks}, is repeated five times.
\label{sec:experiments-evaluation}
We measure the success rate of SFRRT* (\cref{subsec:successrate}) and compare its obstacle avoidance performance (\cref{subsec:orientation-over-time}) against the SpillNot method from \cite{pendulum}. We also evaluate the accuracy of the SFC model on containers that were not part of the training set, which are also used in the video demo (\cref{scenario 2 gif}) to display a real-world chemistry lab procedure (\cref{sec:eval_no-training-set}). Lastly, we present an ablation study that compares modified versions of the SFRRT* (\cref{sec:eval_ablation}). 
\subsection{Implementation Details}
\label{subsec:implement-details}
The performance of SFRRT* depends on the spill constraint enforced during path planning. This constraint restricts the container's orientation by imposing maximum allowable tilt angles to prevent spillage. As described previously, these angles are calculated using \cref{eq:caseI} and \cref{eq:caseII}. \cref{table:tasks}, last column, presents the maximum tilt angles for the containers used in the experiments. To train the spill-free classifier (SFC) model, we deployed it on TensorFlow and optimized it using a binary cross-entropy loss function for 400 epochs with a batch size of 32. SFC achieved an accuracy of 94\%. Detailed training parameters are available on the open-source code.
\begin{table}[H]
\begin{center}
\begin{tabular}{||c | c | c | c | c||} 
 \hline
Experiment & Scenario & Container & Fill Level & $\theta_{max}$\\[0.5ex] 
 \hline
1 & 1 & Wine glass & 80\% & 39$^\circ$\\
\hline
2 & 1 & Wine glass & 30\% & 65$^\circ$\\
\hline
3 & 1 & Flute glass & 80\% & 49$^\circ$\\ 
\hline
4 & 1 & Flute glass & 50\% & 72$^\circ$\\ 
\hline
5 & 2 & Basic glass & 70\% & 35$^\circ$\\
\hline
6 & 2 & Basic glass & 30\% & 62$^\circ$\\
    \hline
\end{tabular}

\caption{\label{table:tasks} \textbf{Experiments designed for assessing SFRRT*}: Each is repeated 5 times using a Panda robotic arm.
}
\end{center}
\end{table}

\vspace{-15pt}
Finally, several key parameters are set in SFTP (\cref{alg:sftp}). The bounds on velocity, acceleration, jerk and are set according to the robotic arm's kinematic constraints \cite{panda}. Moreover, the jerk decrease factor $rate$ is set to 2, meaning when SFTP fails to find a spill-free trajectory, $J_{max}$ will be halved and SFTP will reevaluate for spillage under the updated constraint. This $rate$ selection balances execution time and planning time: a smaller rate would extend the time to find a spill-free trajectory, while a larger rate could lead to unnecessarily slow trajectories due to large jerk constraints. 
\subsection{Success Rate}
\label{subsec:successrate} To measure the success rate of the experiments presented in \cref{table:tasks}, each experiment was repeated five times and the result is presented in \cref{table:spill}. Success is defined as the absence of spillage and environmental collisions during the executed motion. Overall, we achieved a 100\% success rate for all experiments except experiment 5.
For experiment 5, spill-free trajectories were accomplished in four out of five trials. Experiment 5 faced challenges since 
it used a container with a maximum tilt angle of 35$^\circ$, smaller than others, limiting the trajectory planning options and increasing spillage risk. Additionally, the SFC model, with its 94\% accuracy, had difficulty adapting to this constrained scenario, contributing to the lower success rate.
\begin{table}[H]
\begin{center}
\renewcommand{\arraystretch}{1.5}
\begin{tabular}{|c | c | c | c | c | c | c |} 
 \hline
Experiment & 1 &  2 & 3 & 4 & 5 & 6 \\[0.5ex] 
 \hline
Success & $5 / 5$ & $5 / 5$ & $5 / 5$ & $5 / 5$ & $4 / 5$ & $5 / 5$ \\ 
 \hline
\end{tabular}

\caption{\label{table:spill} Success rates for different experiments.}
\end{center}
\end{table}

\subsection{Avoiding Obstacles} 
\label{subsec:orientation-over-time}
\begin{figure}
    \centering
    \includegraphics[width=1\columnwidth]{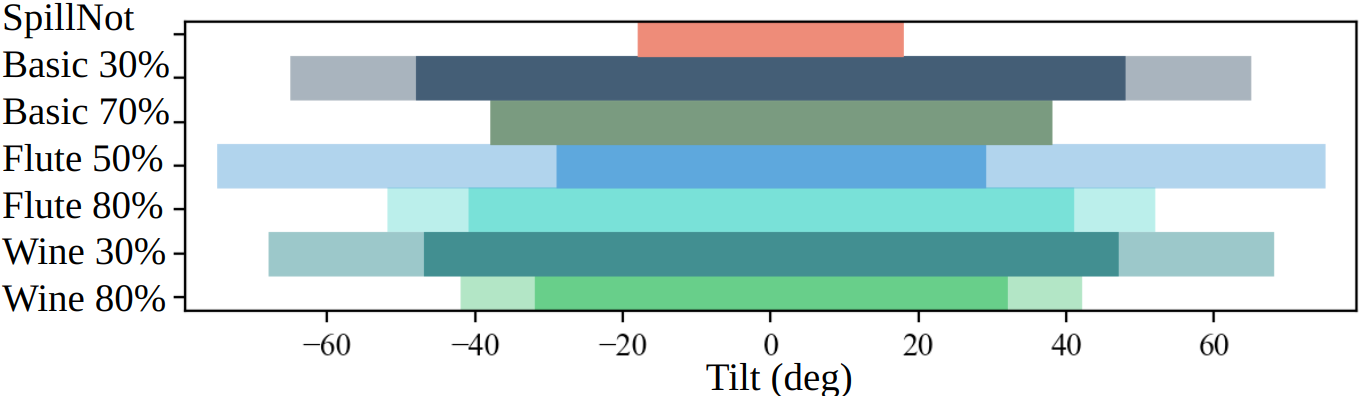}
    \caption{Maximum tilt angles executed by the robot across all 30 trials from \cref{table:tasks}. 
    The Y Axis labels correspond to the experiments, for example, "Basic 30\%" represents experiment 6. The shaded region represents the allowable maximum tilt angle (\cref{table:tasks}, last column). The solid-colored line corresponds to the executed maximum tilt angle. SpillNot's \cite{pendulum} maximum tilt angle is included for comparison.
    }
    \label{fig:tilt_angle_for_diff_tasks}
\end{figure}
A key capability of SFRRT* is its ability to navigate around obstacles in the environment while ensuring the container does not spill. To analyze this ability, we focus on the end effector orientation allowable ranges. In SFRRT*, the container orientation is constrained by the maximum tilt angle of the container, i.e. it can generate spill-free trajectories, so long as the container is not tilted past the maximum tilt angle.
By enabling the robot to tilt the container to higher angles, it can more flexibly maneuver in cluttered spaces, such as a fume hood, which may require tilting chemistry vessels to fit them inside.

 To illustrate this expanded ability, \cref{fig:tilt_angle_for_diff_tasks} shows the maximum possible and maximum utilized tilt angles of all 30 trials of \cref{table:tasks}. 
 In every experiment, SFRRT* utilizes a larger tilt range than the SpillNot \cite{pendulum}.
 Notably, SpillNot's tilt angles are not contingent on the fill level and are always limited to a maximum of 15$^\circ$ to maintain low linearization error (\cite{pendulum}, Fig 4).
 We also chose \cite{pendulum} for comparison because, unlike other methods, it does not rely on sensor-based slosh modeling, and allows translational motion in all axes.

As seen in \cref{fig:tilt_angle_for_diff_tasks}, using SFRRT* the robot achieves a maximum tilt angle of 45$^\circ$ when transporting the basic container with a 30\% fill level (\cref{table:tasks}, experiment 6), surpassing the SpillNot method's 15$^\circ$. The range of orientation for this specific container is 3x larger than \cite{pendulum}. Crucially, the maximum tilt angle for the flute glass with a 70\% fill level is 72$^\circ$, a 4.8x increase in orientation range.

\begin{figure}
    \centering
    \includegraphics[width=1\columnwidth]{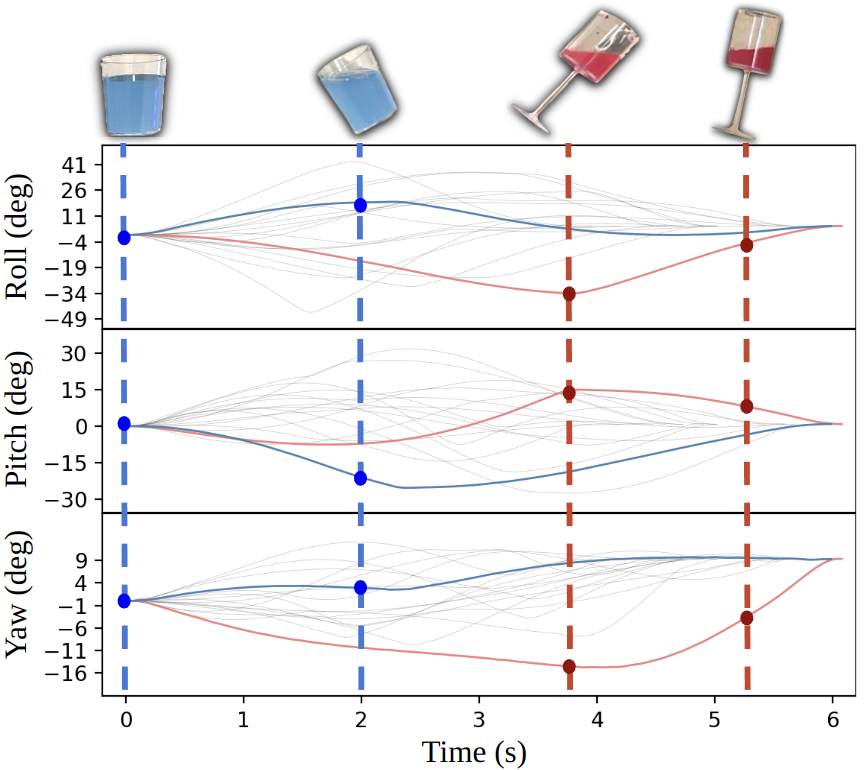}
    \caption{Container orientation over time across all 30 trials from \cref{table:tasks}. The red and blue lines are two trajectories generated for experiment 1 and experiment 6 respectively. Snapshots of the container at specific Euler angles are displayed at the top of the plot. As seen, there are angle changes in each axis of orientation, highlighting the ability of SFRRT* to avoid obstacles.}
    \label{fig:all_trajectories}
\end{figure}

\begin{figure}
\centering
\includegraphics[width=1\columnwidth]{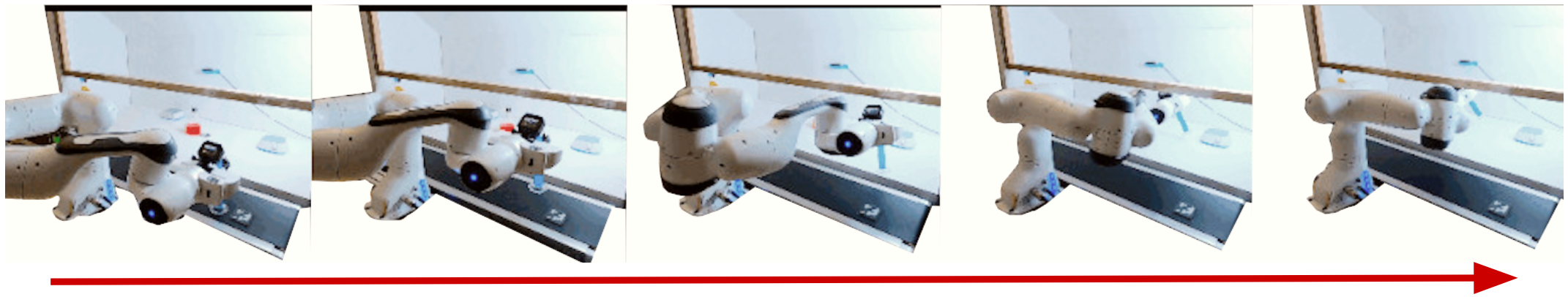}
\caption{The robot transporting a graduated cylinder into the fume hood while avoiding spills and collisions}
\label{scenario 2 gif}
\end{figure}

Furthermore, it is important to enable orientation changes along all axes to avoid obstacles. For example, maintaining a fixed pitch angle throughout the trajectory prevents avoidance of obstacles in certain cluttered environments. SFRRT* can generate trajectories where all three axes of orientation are utilized. \cref{fig:all_trajectories} displays the container orientation over all the 30 trajectories generated for \cref{table:tasks}. As shown, there are changes in Euler angles across all axes, the range for roll, pitch, and yaw angles are $[-49, 41]$, $[-30, 30]$, and $[-16, 9]$ degrees, respectively. In contrast, the SpillNot method imposes a fixed yaw angle throughout the trajectory \cite{pendulum}. 

\subsection{SFC: Generalization Capability and Velocity Adaptation}\label{sec:eval_no-training-set}
The Spill-Free Classifier (SFC) model is trained using 75\% of the 700 trajectories and achieved an accuracy of 94\% for the validation set.
To evaluate the generalization capability of the SFC, its accuracy was tested on novel containers not present in the training dataset. Specifically, a round bottom flask, beaker, and graduated cylinder were used to gather multiple trajectories using a motion capture system (\cref{table:gene}). For each container type, 12 trajectories were gathered with varying fill levels.

As shown in \cref{table:gene}, SFC demonstrates good generalization capability. It achieves strong performance for the beaker and graduated cylinder, with 11 out of 12 data points accurately predicted. However, its accuracy is lower for the round bottom flask, with only 8 out of 12 data points predicted correctly. This is because the SFC model input for container properties only includes details that capture curvy-shaped containers by estimating them to a frustum of a cone. By representing these shapes as frustums of cones, the SFC model ensures a conservative yet safe portrayal of container properties. While this cautious approach may occasionally lead to conservative false negatives, it guarantees the generation of spill-free trajectories.

We also explore a different model architecture for SFC. The current architecture (\cref{fig:models}) passes only the trajectory to the transformer layer before concatenating with container properties. We modified this by concatenating properties at each trajectory step before passing it to the transformer. We call this modified architecture SFC$_m$. SFC$_m$ achieved only 88\% accuracy, lower than the 94\% of SFC, and demonstrated lower accuracy on novel containers than SFC (\cref{table:gene}).
\begin{table}[H]
\begin{center}
\renewcommand{\arraystretch}{1.5}
\begin{tabular}{|c | c | c | c|} 
 \hline
 &
    \begin{minipage}{.2\columnwidth}
        \begin{center}
          \includegraphics[width=.5\columnwidth]{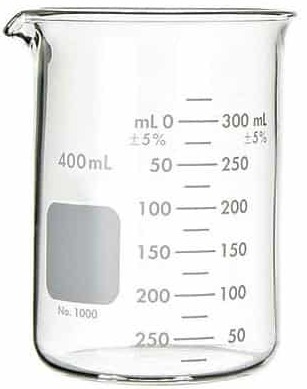}
        Beaker
        \end{center}
    \end{minipage}
 & 
     \begin{minipage}{.2\columnwidth}
     \begin{center}
      \includegraphics[width=.23\columnwidth]{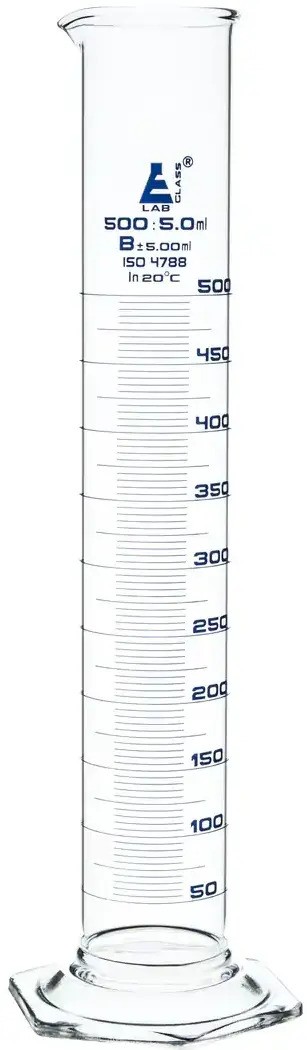}

      Graduated Cylinder
    \end{center}
    \end{minipage}
& 
    \begin{minipage}{.2\columnwidth}
    \begin{center}
      \includegraphics[width=.5\columnwidth]{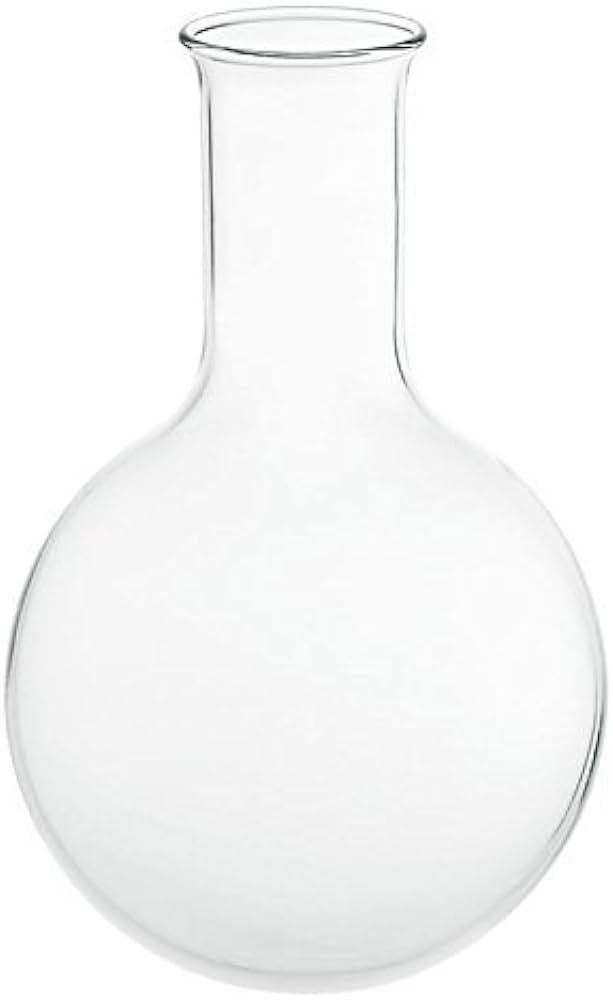}

      Round Bottom Flask
    \end{center}
    \end{minipage}
 \\\hline
\textbf{SFC} & \textbf{$11 / 12$} & \textbf{$11 / 12$} & \textbf{$8 / 12$} \\\hline
SFC$_{m}$& $8 / 12$ & $9 / 12$ & $8 / 12$ \\\hline
\end{tabular}
\caption{\label{table:gene} Model accuracy using unseen containers. SFC demonstrates robust performance in predicting outcomes with previously unseen containers, outperforming SFC$_m$.
}
\end{center}
\end{table}

\begin{figure}
    \centering
    \includegraphics[width=0.8\columnwidth]{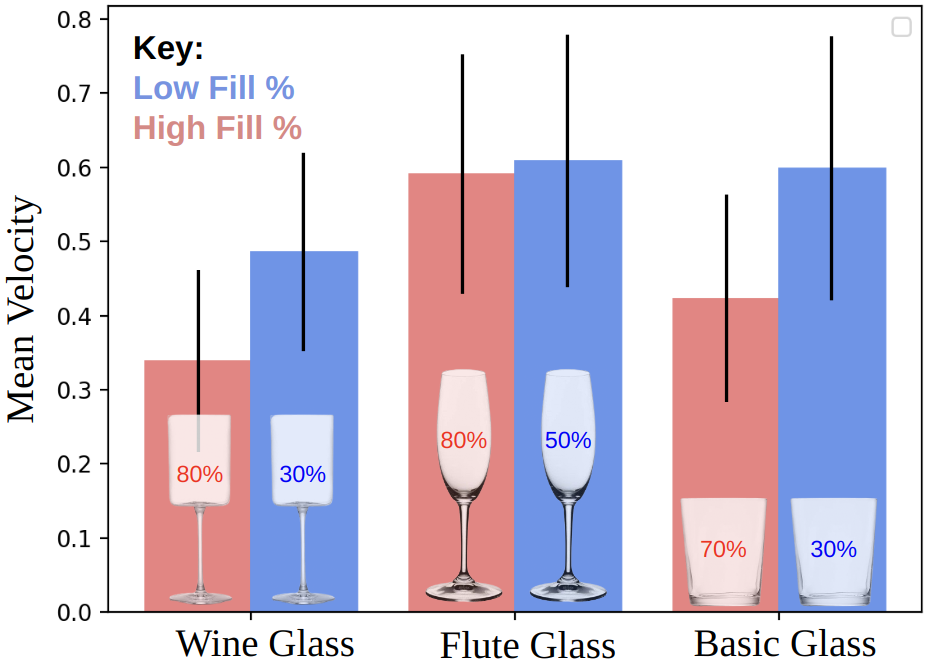}
    \caption{\textbf{Executed Mean Velocity for Containers from \cref{table:tasks} Using SFRRT*} As depicted, SFRRT* performs faster trajectories for containers with lower liquid fill levels.
    }
    \label{fig:mean-vel}
\end{figure}

Lastly, SFC is able to take advantage of the container shape to generate trajectories with different velocities. As seen in \cref{fig:mean-vel}, considering the same containers, with different fill levels we see that it consistently generates higher velocities for containers with lower fill levels, which are less likely to spill. Moreover, when comparing different containers in the same scenario (as in \cref{table:tasks}), those with higher tilt angles consistently achieve higher velocities. For example, in the comparison between the wine glass and the flute glass, the wine glass consistently achieves lower mean velocities. This is attributed to the geometry of the containers. As shown in \cref{table:tasks}, the wine glass has a lower maximum tilt angle compared to the flute glass, indicating smaller room to maneuver and a higher susceptibility to spillage. Conversely, the flute glass, with its higher maximum tilt angle consistently achieves higher mean velocities.
\subsection{Ablation Study}\label{sec:eval_ablation}
In this section, two key elements of SFRRT* are modified to evaluate their impact on performance: the informed sampler and the spill-free classifier (SFC) model.
\begin{table}[H]
\begin{center}
\renewcommand{\arraystretch}{1.5}
\begin{tabular}{|c | c | c| c|} 
 \hline
 Task & \textbf{SFRRT*} & SFRRT*$_{u}$ & SFRRT*$_{r}$ \\\hline
  1 & \textbf{$5 / 5$} & $0 / 5$ & $1 / 5$ \\\hline 
  2 & \textbf{$5 / 5$} & $1 / 5$ & $1 / 5$ \\\hline
\end{tabular}

\caption{\label{table:ablation} \textbf{Success Rate Comparison:} 
SFRRT*$_{u}$ uses a uniform sampler, while SFRRT*$_{r}$ skips the SFC spillage checks and uses random sampling for jerk constraints in time parameterization. Overall, SFRRT* outperforms all variants.
}
\end{center}
\end{table}

\textit{Informed Sampler Variation: SFRRT*$_{u}$.}
To avoid sampling states such as an upside-down container, or a container tilted beyond its capacity, the informed sampler samples states within the maximum allowable tilt angles for the container (\cref{table:tasks}, last column). To measure its impact, we replace it with a uniform sampler, creating SFRRT*$_{u}$. Then, experiments 1 and 2 are executed 5 times on the robot, and the success rate is shown in \cref{table:ablation}. The success rate drops significantly since the path generation lacks state sampling constraints, as a result, the time parameterization algorithm struggles to create a spill-free trajectory using this path.

\textit{Spill-Free Classifier Variation: SFRRT*$_{r}$.}
In SFRRT*, the SFC model assists with time parameterization by selecting the jerk constraints to ensure a spill-free trajectory. To evaluate the impact of SFC, we introduce SFRRT*$_{r}$ by replacing the SFC model with a random jerk selection. More specifically, a random jerk is selected out of all the jerks that were ever chosen by SFC. The path is then parameterized using that jerk. \cref{table:ablation} shows the success rate for experiments 1 and 2. The observed drop in success rate is due to the random selection of jerk constraints without considering the path itself, in contrast to using the SFC model which takes the path into account when choosing the jerk constraints.

\section{Conclusion and Future Work}\label{sec:conclusion}
This work presents SFRRT*, a sampling-based motion planner for the spill-free transport of open-top, liquid-filled containers, particularly in cluttered scenes. SFRRT* distinguishes itself by utilizing implicitly learned spillage dynamics via a transformer-based machine-learning approach. This outperforms existing approaches by increasing the success rate in cluttered scenes. Specifically, SFRRT* enables motions across all axes of end effector orientations using the learned model instead of simplified liquid models. It also permits the container to tilt to its maximum tilt angle, thus enabling flexible maneuvering to avoid obstacles. It also does not need visual monitoring of the liquid, which could get obstructed in cluttered scenes. 

However, like many machine learning models, our approach faces a common limitation— the need for a substantial amount of training data. Moreover, containers are approximated as frustums of cones to estimate the maximum tilt angles. This could lead to conservative estimates for containers with more complex curvy shapes. An alternative would be training a model on image inputs to predict tilt angles more precisely. Lastly, circular upside-down motions (loop-de-loop) can avoid spilling if executed with adequate speed. Still, SFRRT* cannot support generating them because it constrains container orientations to be less than the maximum tilt angle. This limitation stems from the fact that SFC is trained on data collected from humans, who typically exhibit caution when handling liquids. Consequently, the model lacks training data to classify loop-the-loop motions as spill-free.

For future work, we are interested in improving the Spill-Free Classifier model to estimate spillage probability rather than solely classifying it, thus allowing for a more precise time-parameterized trajectory. Furthermore, the path planning phase currently only constrains the orientation range within the maximum tilt angle. However, if the sampled orientation in the resulting path approaches the maximum tilt angle, the generated trajectory will be very slow. To address this, we plan to analyze how the sampled orientation affects velocity, acceleration, and jerk during path planning. We aim to generate a time-optimal parameterized path by considering these kinematic factors in the path-planning phase.
Moreover, our approach is versatile and can be expanded to a variety of applications. We plan to extend our method to include spillage dynamics for granular materials such as sand and sugar, to transport a tablespoon of these materials without spillage.
Furthermore, our method is adaptable to different path planners and not solely dependent on RRT*. Moving forward, we plan to evaluate our approach using various planning algorithms and compare success rates and planning times.
Finally, to improve our SFC accuracy and SFRRT* success rate further, we aim to optimize our data collection process by exploring automation through Computational Fluid Dynamics (CFD) frameworks.

\AtNextBibliography{\small}
\printbibliography
\end{document}